\documentclass[10pt,twocolumn,letterpaper]{article}

\usepackage[pagenumbers]{cvpr} %
\usepackage{booktabs}
\usepackage{multirow}
\usepackage{graphicx}
\usepackage{pdflscape}
\usepackage{mathabx}
\usepackage{txfonts}
\usepackage{placeins}
\usepackage[normalem]{ulem}
\useunder{\uline}{\ul}{}
\usepackage[mode=match]{siunitx}

\usepackage[dvipsnames]{xcolor}

\def\eqref#1{Eq.~(\ref{#1})}

\usepackage{array}
\newcolumntype{L}[1]{>{\raggedright\let\newline\\\arraybackslash\hspace{0pt}}m{#1}}
\newcolumntype{C}[1]{>{\centering\let\newline\\\arraybackslash\hspace{0pt}}m{#1}}
\newcolumntype{R}[1]{>{\raggedleft\let\newline\\\arraybackslash\hspace{0pt}}m{#1}}

\usepackage{colortbl}
\definecolor{TableGray}{gray}{0.9}

\definecolor{cvprblue}{rgb}{0.21,0.49,0.74}
\usepackage[pagebackref,breaklinks,colorlinks,citecolor=cvprblue]{hyperref}

\title{The Bare Necessities: \\ Designing Simple, Effective Open-Vocabulary Scene Graphs }

\author{
Christina Kassab$^{1}$ \qquad Matías Mattamala $^{1}$ \qquad Sacha Morin$^{2}$\qquad Martin Büchner$^{3}$\\ Abhinav Valada$^{3}$ \qquad Liam Paull$^{2}$ \qquad Maurice Fallon$^{1}$  \vspace{0.05cm}\\
{\small $^{1}$ University of Oxford \quad $^{2}$ Université de Montréal \quad $^{3}$ University of Freiburg }
}

\begin{document}

\twocolumn[{%
\maketitle
\vspace{-3.4em}
\begin{center}
    \centering
	\includegraphics[width=\textwidth]{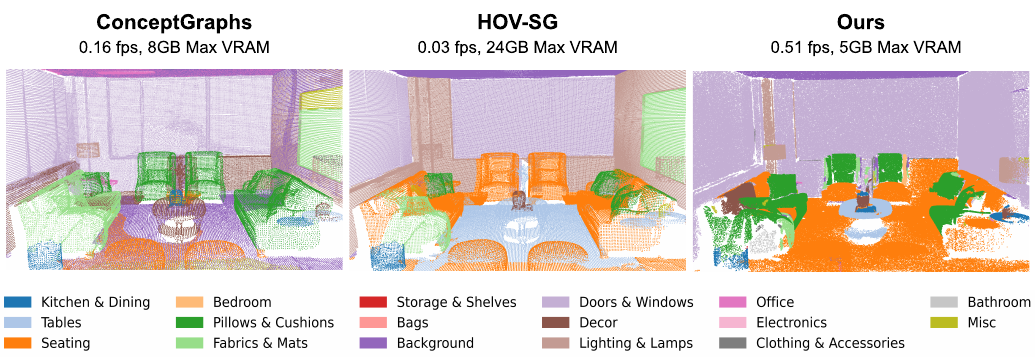}
     \captionof{figure}{\small \textbf{3D open-vocabulary segmentations and processing times from three different methods: ConceptGraphs (left), HOVSG (middle) and, our proposed system (right)}. In this work, we analyze open-vocabulary scene graph methods and use our findings to develop a minimal system which achieves comparable object classification accuracy at a fraction of the computational cost of existing methods. We show object labels grouped into broader categories for easy visualization, for more details see Supp. Sec~\ref{sec:final_results}.
    	}
\end{center}%
}]

\begin{abstract}
3D open-vocabulary scene graph methods are a promising map representation for embodied agents, however many current approaches are computationally expensive. In this paper, we reexamine the critical design choices established in previous works to optimize both efficiency and performance. We propose a general scene graph framework and conduct three studies that focus on image pre-processing, feature fusion, and feature selection. Our findings reveal that commonly used image pre-processing techniques provide minimal performance improvement while tripling computation (on a per object view basis). We also show that averaging feature labels across different views significantly degrades performance. We study alternative feature selection strategies that enhance performance without adding unnecessary computational costs. Based on our findings, we introduce a computationally balanced approach for 3D point cloud segmentation with per-object features. The approach matches state-of-the-art classification accuracy while achieving a threefold reduction in computation.
\end{abstract}
    
\section{Introduction}
\label{sec:intro}

Scene understanding is an important capability for embodied agents. Humans can simultaneously perceive a scene at multiple levels of abstraction. We consider global properties such as the ground, walls, and ceilings; local properties such as objects and their interrelationships; and affordances associated with the space, such as cooking, dining, studying, etc.~\cite{Clemens2023}. For an embodied agent, an ideal scene representation should be constructed and updated in \textit{real-time}, using onboard sensors and compute. It should be \textit{scalable} to accommodate large-scale, multi-floor environments, \textit{generalizable} to support new types of objects, and sufficiently \textit{versatile} to support a variety of downstream tasks such as navigation or manipulation~\cite{davison2018futuremappingcomputationalstructurespatial}.

An emerging approach for this is 3D scene graphs~\cite{hughes2024foundations, ConceptGraphs2023, hovsg2024, koch2024open3dsg, takmaz2023openmask3d, Maggio2024Clio, jiang2024roboexpactionconditionedscenegraph}, which provide a structured representation that seeks to implement this conceptualization. Scene graphs are often organized hierarchically by incorporating multiple levels of semantic abstraction---from rooms to individual objects. To enable the representation to support a large number of object categories, many recent approaches now use open-vocabulary models such as CLIP~\cite{clip2021} and compare against text encodings to label objects and scenes~\cite{ConceptGraphs2023, hovsg2024, koch2024open3dsg, takmaz2023openmask3d, Maggio2024Clio}. 

While this progress is promising, building scalable scene graphs in real-time remains a significant challenge. Current approaches such as ConceptGraphs~\cite{ConceptGraphs2023} and HOV-SG~\cite{hovsg2024} rely on large models such as SAM~\cite{kirillov2023segany}, GPT~\cite{openai2024gpt4technicalreport}, or LLaVa~\cite{liu2023llava}, which are prohibitively expensive for real-time use. These models typically process images at multiple scales to capture varying levels of context around each object. They then pool these individual observations to classify the object into a category. However, recent studies~\cite{fan2024, song2024} indicate that CLIP's performance is highly dependent on the viewpoint from which an object is observed which raises concerns about the effectiveness of naively averaging multi-view features. Research in the field has also not fully explored whether other commonly used system components significantly improve the performance of scene graph methods. 

In this work, we revisit these assumptions by presenting an analysis of the different design decisions taken by recent scene graph approaches. Our aim is to provide insights on how to maximize performance while targeting the efficiency needed for real-time systems. To achieve this, we first define a typical framework for scene graph methods and analyze their constituent modules. We focus on image pre-processing, which we identify as one of the main bottlenecks for real-time operation. We further explore how information from multiple views, such as from CLIP features, is fused.

Using the findings of our study, we then develop a minimal demonstrative open-vocabulary system. We particularly focus on classifying objects and therefore producing the \textit{graph nodes}. Our method takes as input a set of RGB-D images and camera poses, which can be produced by an online 3D reconstruction system. Its output is a segmented point cloud of a scene with a single CLIP feature assigned to each segment. Our method achieves performance comparable to state-of-the-art methods such as ConceptGraphs~\cite{ConceptGraphs2023} or HOV-SG~\cite{hovsg2024} in 3D segmentation tasks (in terms of metrics such as mIOU and mAcc)  while being more than three times faster. Our findings show that by better focusing computation on key components, a simpler and more effective scene graph method is achievable.

The contributions of this paper are as follows: 
\begin{itemize}
    \item We analyze the structure of current scene graph systems and propose a general framework to describe their operation. We then use this framework to analyze the key system components in three experimental studies.
    \item Our first study on pre-processing strategies demonstrates that averaging features from scaled image crops and SAM masks provides minimal improvement in object classification accuracy while increasing computation time more than threefold per object view.
    \item Our second study on strategies for open-vocabulary feature fusion shows that CLIP produces highly variable outputs from different views of the same object and that averaging across multiple views does not improve but actually reduces classification accuracy.
    \item Our third study on feature selection examines alternative strategies for choosing representative features from multi-view data. Our results show that using entropy to select optimal views enhances classification performance without increasing computation time.
    \item Finally, we integrate these findings into the design of an open-vocabulary system. The system achieves comparable performance to state-of-the-art methods while achieving a threefold reduction in computation.
\end{itemize}

\section{Related Work}
\label{sec:related_work}

\begin{figure*}[ht]
    \centering
    \includegraphics[width=\textwidth]{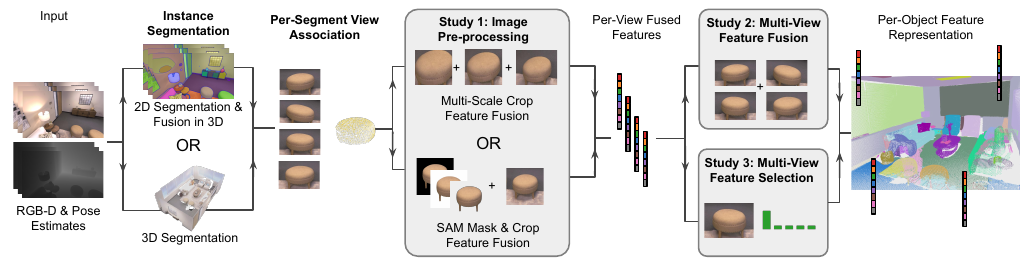}
    \caption{\small{\textbf{Typical framework used by open-vocabulary scene graph methods.} The pipeline includes: 1) Input RGB-D images and poses. 2) Perform 3D instance segmentation via 2D segmentation or point cloud accumulation. 3) Select unobstructed views based on visible points. 4) Scale and fuse features or apply a SAM mask. 5) Fuse features across views for final object labeling. Highlighted sections are explored in Section~\ref{sec:study}.}}
    \label{fig:ov_graph_pipeline}
\end{figure*}

\textbf{3D Semantic and Instance Segmentation.} 3D segmentation systems fall into two main categories: semantic segmentation, which assigns a label to each point in a 3D scene~\cite{Anand2013, Armeni_2016_CVPR, Choy2019, Graham2018, hu2021vmnet, hua-pointwise-cvpr18, Weder2024}, and instance segmentation, which further differentiates individual instances within the same object class~\cite{schult2023mask3dmasktransformer3d, engelmann20203dmpamultiproposalaggregation, Han_2020_CVPR, zhou2023uni3dexploringunified3d, jiang2020pointgroup, Vu2022SoftGroupF3}. Other methods are purely geometry-based and perform segmentation without the need for prior training on specific classes. A classic example of this is region growing~\cite{Adams1994}. More recent approaches such as SAM3D~\cite{yang2023sam3d} either draw inspiration from 2D methods, such as SAM~\cite{kirillov2023segany,Ravi2024SAM2} or leverage geometry and color priors as in SA3DIP~\cite{yin2024sai3dsegmentinstance3d}.

\textbf{3D Scene Graphs.} Differently from 3D segmentation, 3D scene graphs do not necessarily aim to label every point in a scene. Instead, they build a high-level representation by structuring objects and their relationships in a graph format. Early scene graph works such as Hydra~\cite{Hughes2022} and S-Graphs+~\cite{bavle2023sgraphsrealtimelocalizationmapping} construct multi-layer graph representations, with nodes representing places or objects and edges representing semantic relationships. Semantics are often obtained by using closed-vocabulary pretrained networks such as HR-Net~\cite{SunXLW19}, and room segmentations are obtained using Graph Neural Networks (GNNs) or geometry. 

Recent works, such as Open3DSG~\cite{koch2024open3dsg}, ConceptGraphs~\cite{ConceptGraphs2023}, and HOVSG~\cite{hovsg2024}, construct 3D representations using nodes and edges while incorporating open-vocabulary models like LLaVa~\cite{liu2023llava}, GPT~\cite{openai2024gpt4technicalreport}, and CLIP~\cite{clip2021} to add semantics. OpenMask3D~\cite{takmaz2023openmask3d} follows a similar framework but does not have explicit edge relationships. These methods typically follow a similar framework: first, segments are obtained in 2D or 3D~\cite{openai2024gpt4technicalreport}, then projected into 3D and merged~\cite{ConceptGraphs2023, hovsg2024}. Each segment is associated with multi-view images that are pre-processed (e.g., through scaling or masking using segmentation networks like SAM). Features are then extracted from the pre-processed images from image-text models such as CLIP. Features can be averaged across scales~\cite{takmaz2023openmask3d} or across both scales and masks~\cite{hovsg2024}. The multi-view features are then typically averaged~\cite{takmaz2023openmask3d, ConceptGraphs2023} or filtered with DBSCAN to perform majority-vote classification, yielding per-object features~\cite{hovsg2024}. The objects are treated as nodes in a graph, where edges represent relationships between objects. Edge labels can be determined through geometric relationships or through an LLM as in~\cite{ConceptGraphs2023}. The objects can also be attached to higher-level layers in the graph which encode concepts such as rooms and floors~\cite{hovsg2024}.

Although these methods offer comprehensive semantic representations suited for tasks such as navigation and object retrieval, they are often limited by their lack of real-time capabilities and scalability. Open vocabulary works that achieve real-time performance include Clio~\cite{Maggio2024Clio} and LEXIS~\cite{Kassab2024}. Both have limitations: Clio relies on a task-specific representation from an input prompt list, while LEXIS is limited to individual room segmentation. 

\textbf{Multi-View CLIP.} A particular focus of this work is analyzing how the performance of CLIP varies with different views and the adverse effect that averaging features across multiple views can have on object classification accuracy. Previous studies have proposed techniques to mitigate this variance. Song~\textit{et~al.}~\cite{song2024} integrates a self-attention mechanism for robust multi-view feature aggregation and a navigation policy to select optimal views for objects. Fan~\textit{et~al.}~\cite{fan2024} uses the entropy of the detected classes'  distribution as a confidence metric, followed by an LLM-driven prompt system that refines classification.

Other methods fine-tune CLIP using 3D data~\cite{lee2024duoduo, zhou2023uni3dexploringunified3d, liu2023openshape}. Duoduo CLIP~\cite{lee2024duoduo} uses contrastive learning in addition to cross-view attention to fuse information across multiple views of an object and boost performance. Uni3D~\cite{zhou2023uni3dexploringunified3d} is trained to align 3D point cloud features with the image-text features from CLIP. In contrast, OpenShape~\cite{liu2023openshape}, uses a multi-modal contrastive framework to learn joint representations of text, image and point clouds.  We exclude these methods from our analysis as they are resource-intensive and not well-suited for our goal of developing real-time applications. Instead, we focus on strategies that directly fuse or select CLIP features from multiple views, which we examine in detail in the following study.

\section{A Study of Open-Vocabulary Scene Graphs}
\label{sec:study}

\begin{figure*}[ht]
    \centering
    \includegraphics[width=\textwidth]{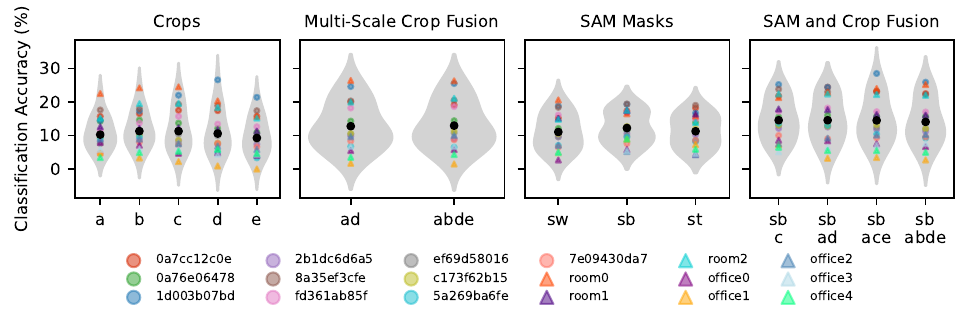}
    \caption{\small{\textbf{Classification accuracy using different pre-processed images}, including crops and masks generated by SAM2. $\ovoid$ corresponds to ScanNet++ scenes and $\bigtriangleup$ to Replica scenes. $\medbullet$ indicates the mean. We show crops at different scale factors, crops fused over multiple scales, SAM masks with various backgrounds, and SAM masks fused with crops. The letters denote either the scale or the SAM mask: a = 1.0, b = 1.2, c = 1.5, d = 1.8, e = 2.0, sw = SAM mask with a white background, sb = SAM mask with a black ground and st = SAM mask with a transparent background. Any combinations indicate fusions. The results indicate that the various pre-processing methods do not improve performance significantly.} }
    \label{fig:preprocessing}
\end{figure*}

We begin with an analysis of the structure of current open-vocabulary scene graph methods. 

\noindent\textbf{Scene graph framework.} We identify that most existing approaches follow a similar framework, illustrated in Fig.~\ref{fig:ov_graph_pipeline}. The input to these methods is a set of RGB-D images and their corresponding poses. The output is a segmented point cloud with open-vocabulary features per object.

To achieve this, a 3D instance segmentation is first generated using one of two main approaches. The first approach segments the 2D RGB images using an instance segmentation method such as SAM. The segments are then projected into 3D using depth images and fused over time. The other main approach performs 3D instance segmentation on a reconstructed point cloud or mesh created from RGB-D images and poses, using a network such as Mask3D~\cite{schult2023mask3dmasktransformer3d}. Each segment can then be associated with a set of images, typically only retaining the ``best'' views in which the segment is most visible. These images are pre-processed by scaling or masking. By adjusting the crop scale, we can provide context for understanding the object within the scene. On the other hand, masking highlights the main object of interest while minimizing irrelevant details~\cite{shtedritski2023does}. For each segment viewpoint, CLIP features are extracted from these images and averaged to obtain a single feature per viewpoint. 

Finally, these viewpoint features are fused, often by averaging, to create a final per-segment feature. Each segment and its associated feature are represented as nodes; object relationships (such as ``on top of'' or ``beside'') are expressed as edges, which can be captured by an LLM~\cite{ConceptGraphs2023}. The objects can also be associated with higher-level concepts such as rooms or floors~\cite{hovsg2024}.

\noindent\textbf{What we study.} We study three main components of this framework, as highlighted in Fig.~\ref{fig:ov_graph_pipeline}: The first study examines the impact of image pre-processing, particularly scaling and the use of masks. The second study evaluates the effectiveness of averaging for multi-view feature fusion. The third study explores an alternative to fusing features based on selecting a representative feature per object.

\noindent\textbf{What we did not study.} We exclude an evaluation of the segmentation step and, instead, conduct our evaluation using ground truth object segments. The problem of 3D segmentation has been widely addressed in previous works, including both fusion of 2D segments in 3D~\cite{Gan_2020, WANG2022104185, rosinol2020kimeraopensourcelibraryrealtime} and 3D instance segmentation networks~\cite{isprs-archives-XLII-2-W3-339-2017, Muhammad_Yasir_2022}. We also do not analyze approaches to build the graph and edge relationships and instead focus on generating the graph nodes.

\noindent\textbf{Datasets.} We conduct this evaluation using the ground-truth segmentation from two publicly available datasets: ScanNet++ and Replica. From ScanNet++, we use 10 representative scenes, including student bedrooms, a lab, offices, apartments, and classroom scenes. From Replica, we use \textit{office0-office4} and \textit{room0-room2} as used in other works such as HOV-SG and ConceptGraphs. In our first study, we use five different scales for cropping: 1.0, 1.2, 1.5, 1.8, and 2.0. Additionally, we generate a SAM mask for each object, setting the background to be white, black or transparent (alpha = 0). We use SAM2~\cite{Ravi2024SAM2} for this evaluation and run all experiments on a mid-range laptop (for more implementation details see Supp. Sec~\ref{sec:compute}, \ref{sec:datasets}, \ref{sec:crop_generation}, and \ref{sec:classification_accuracy}).

\noindent\textbf{Evaluation Prompt List.} We use an extended prompt list comprised of the labels from all scenes in ScanNet++ and Replica to evaluate the produced open-vocabulary representations (1687 labels in total). We refer to this prompt list as the \textit{evaluation prompt list}. Current open-vocabulary scene graph works rely on the ground-truth list of scene labels for the prompt list for evaluation, creating an overly controlled assessment setting. In real-world scenarios, queries for a single object can take many forms. For instance, when referring to a sofa, a user might ask for a chair, couch, settee, or a place to sit. Restricting the prompt list to ground-truth labels for a specific scene fails to capture this variability, ignoring the broader range of potential user queries that CLIP might handle. Therefore, in this analysis, we aim to examine how strategies such as feature fusion and selection are impacted by this more realistic evaluation approach.

\subsection{Study 1: Image Pre-processing}
\label{subsec:pre-processing}

We first evaluate image pre-processing techniques that generate curated views with only the object of interest visible, aiming to improve the quality of CLIP features for each view. Prior work uses two main methods: \textit{cropping}, which uses the object's 3D position to crop the image around it with optimal scaling for context, and \textit{masking}, which segments the object using segmentation models such as SAM. Mixed methods also exist, such as multi-scale cropping to average features across scales~\cite{takmaz2023openmask3d} or averaging features produced from crops and SAM masks~\cite{hovsg2024}. We identify this component as the major computational bottleneck for current methods, as it requires extracting CLIP features and SAM masks multiple times per object view which increases processing time and resource demands.

We analyze the effect of these differing methods on object classification accuracy, with the results shown in Fig.~\ref{fig:preprocessing} (further results are presented in Supp. Sec~\ref{sec:preproc_table}). While these methods have been widely used in the past, our results indicate that object classification accuracy remains largely unaffected by variations in input crop size, multi-scale crop fusion, SAM masks, or their combinations, especially when larger prompt lists are employed.

To optimize for efficiency, a single crop scaled to a factor of 1.5 is the most effective choice, achieving a classification accuracy of 11\%. We also find that fusing features across multiple scales provides minimal performance boost while increasing inference time with each additional rescaling.

Applying a mask alone does not enhance classification accuracy compared to simply scaling the crop. Fusing SAM masks with crops at different scales yields a 3\% performance increase, though it comes with a three-fold increase in computation time per view. Extracting CLIP features for a single crop takes \SI{78.1}{\milli\second} while fusing a crop with SAM masks extends this to \SI{314}{\milli\second}. Consistent with findings in~\cite{hovsg2024}, we observe that if masking is desired, a black background produces the best results.

The variance in classification accuracy across the test scenes is up to 8\% (roughly 3 or 4 objects misclassified per scene). We attribute it to some scenes being highly cluttered, with objects being partially obstructed. As we are interested in building a lightweight open-vocabulary solution we use the single crop with a scale factor of 1.5 for all our further analysis.

\subsection{Study 2: Multi-View Feature Fusion}
\label{subsec:feature-fusion}

Since pre-processing images does not yield significant performance gains, we focus on other components of the pipeline---specifically, feature fusion---to explore potential improvements without increasing processing time. After extracting representative CLIP features for each view of the object of interest, these features are fused into a single representative feature \emph{per object}. A common method for multi-view feature fusion is averaging~\cite{ConceptGraphs2023, takmaz2023openmask3d}, though alternative approaches use clustering methods, such as DBSCAN, to obtain the dominant feature~\cite{hovsg2024}. We disregard this method due to the computational complexity of clustering CLIP features with a large number of dimensions.

\begin{table}[t]
\centering
\caption{\small{\textbf{Classification accuracy of multi-view feature fusion vs. CLIP upper bound performance}. Averaging across views or selecting the class mode of the predicted classes achieves significantly poorer performance than the best achievable upper bound due to the sensitivity of CLIP to the viewpoint.}}
\label{tab:averaging}
\footnotesize %
\begin{tabular}{@{}lc|cc@{}}
\toprule
          & \multicolumn{1}{l}{\textbf{Upper-Bound}} & \multicolumn{1}{l}{\textbf{Average}} & \multicolumn{1}{l}{\textbf{Mode}} \\ \midrule
ScanNet++ & 31.2                            & 13.0                          & 13.6                     \\
Replica   & 25.1                            & 11.8                        & 11.1                     \\ \bottomrule
\end{tabular}
\vspace{-5pt}
\end{table}

We suggest that \emph{the upper bound that any fusion strategy could achieve is to match the best classification performance achieved by any feature independently}. We define the \emph{upper bound} as: if any of the available views produce a label that matches the ground truth, the object is deemed to be classified correctly. If there exists a feature from a view that predicts the correct label, this performance will necessarily decrease when averaging unless all the features are in strong agreement---which would suggest a uni-modal distribution in feature space.  We compare two alternative strategies to this upper bound: the average of the multi-view features or the mode label predicted by the features.

We report the classification performance in Tab.~\ref{tab:averaging}. We observed that no apparent improvements are obtained when aggregating different views in feature space, which is in agreement with previous works~\cite{fan2024, song2024}. Both the averaging and the mode strategies achieve similar performance which falls significantly below the upper bound. 

\begin{figure}[b]
    \centering
    \includegraphics[width=1\columnwidth, clip, trim={0 0 0 0}]{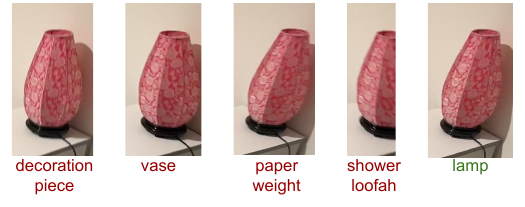}
    \caption{\small \textbf{Example of the multi-view variance of CLIP outputs}. We show various views of a lamp from ScanNet++ scene 0a7cc12c0e, with crops scaled to 1.5x of the original bounding box. The number of visible points per viewpoint is similar in each view but the output varies greatly indicating that visibility is not the optimal method for choosing ``best'' views.} 
    \label{fig:mv_variance}
\end{figure}

Figure~\ref{fig:mv_variance} presents an example of the outputs generated by CLIP for different viewpoints of an object. As illustrated, the output classification can vary significantly between views, regardless of the images being very similar. This indicates that selecting ``optimal'' views based on visibility alone does not ensure the best viewpoint or image quality for maximizing CLIP performance. In many cases, only one or a few viewpoints from the provided set yield the correct response from CLIP. This variation can be attributed to several factors, including the viewpoint, the object's visibility, and the overall quality of the image. Therefore, we cannot assume that strong agreement in multi-view labels will yield a representative label, and so other strategies must be explored (for further examples of CLIP's multi-view variance see Supp. Sec~\ref{sec:multi-view-variance-examples}).

\subsection{Study 3: Multi-View Feature Selection}
\label{subsec:feature-selection}

\begin{figure}
    \centering
    \includegraphics[width=1\columnwidth, clip, trim={0 0 0 0}]{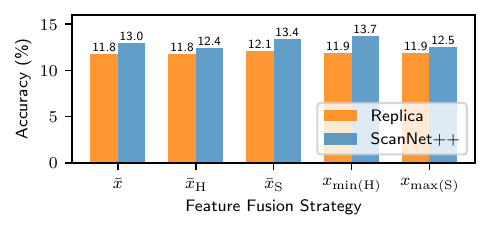}
    \caption{\small \textbf{Object classification accuracy using different multi-view feature fusion and selection strategies}. $\bar{x}$, refers to averaging the multi-view features, $\bar{x}_{\footnotesize{\mathrm{H}}}$ is a weighted average where the weights are the entropy values, $\bar{x}_{\footnotesize{\mathrm{S}}}$ is the weighted average based off of the score, $x_{\footnotesize{\mathrm{min(H)}}}$ is selecting the feature with the lowest entropy, and $x_{\footnotesize{\mathrm{max(S)}}}$ is selecting the feature with the highest score.} 
    \label{fig:mv_feature_fusion}
\end{figure}

Our previous study showed that fusion strategies not only fail to enhance classification performance but also fall short of the upper bound. We propose that more effective feature selection could achieve performance gains. Thus, this study focuses on exploring multi-view feature selection strategies.

When CLIP image features are compared to text encodings, the resulting probability distribution across the label space---referred to here as inter-concept similarity, $s^{\text{concept}}_t$---can be analyzed to assess uncertainty. Previous works have explored the shape of this distribution, hypothesizing that highly skewed distributions indicate low ambiguity and, therefore, high confidence in the features~\cite{song2024, fan2024}. These works have shown that by selecting views based on the Shannon entropy (a metric that quantifies the uncertainty or unpredictability of a set of outcomes) object classification performance can be improved.

\noindent\textbf{An Initial Look at Entropy.} We first study how to select features based on the Shannon entropy. We calculate it from the inter-concept similarity histogram, using the full list of 1687 labels as the text prompt. We compare different entropy selection strategies (simple feature averaging, weighted average, minimum entropy) with a strategy that chooses features by comparing the maximum score of the inter-concept similarity.

\begin{figure}[t]
    \centering
    \includegraphics[width=1\columnwidth, clip, trim={0 0 0 0}]{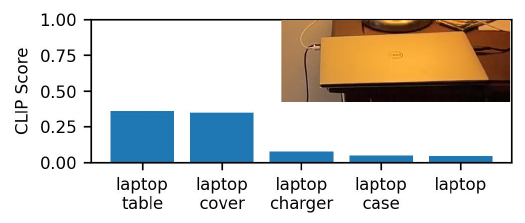}
    \caption{\small \textbf{Histogram showing the CLIP response to an image of a laptop}. The top 5 results all contain variations of laptop-related labels, resulting in a low entropy score and high uncertainty even though the labels are semantically similar. The image was taken from ScanNet++ scene 0a76e06478.} 
    \label{fig:histogram}
\end{figure}

As shown in Fig~\ref{fig:mv_feature_fusion}, neither weighting the features based on entropy nor selecting the feature with the lowest entropy results in any significant improvement in classification accuracy. Similarly, using CLIP's confidence score yields comparable results. These findings stem from the high sensitivity of entropy to the prompt list used for calculating inter-concept similarity. Previous studies often rely on limited prompt lists (using about 50 categories), which may fail to capture all objects in the environment, leading CLIP to produce positive responses even for unrepresented categories. In contrast, a larger prompt list, as used in this analysis, can inflate entropy by responding equally to closely related terms (e.g., ``laptop case'' and ``laptop bag''), see Fig.~\ref{fig:histogram}. 

\begin{figure}[t]
    \centering
    \includegraphics[width=1\columnwidth, clip, trim={0 0 0 0}]{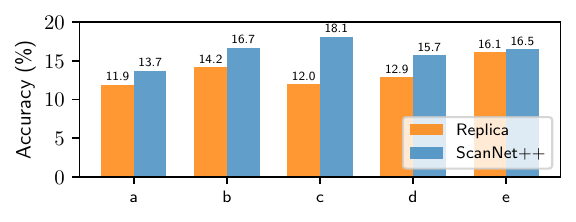}
    \caption{\small \textbf{Impact of prompt list on entropy in object classification feature selection}. In all cases, we select the feature with the lowest entropy. The letters refer to different prompt lists: a is the full list of 1687 labels, b is the same list with duplicates/similar labels removed, c is the first 500 labels from ScanNet++, d is a random set of 500 objects, and e is a list of 50 Replica labels. } 
    \label{fig:mv_feature_fusion_improved}
\end{figure} 

 \begin{figure*}[ht]
    \centering
    \includegraphics[width=\textwidth]{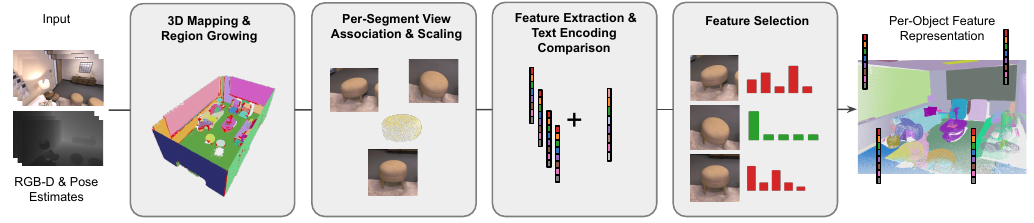}
    \caption{\small \textbf{System Overview.} We process input RGB-D images and camera poses to generate point clouds and perform region growing for 3D segmentation. For each segment, we crop images around the object's bounding box, extract CLIP features, and compute the inter-concept similarity. Finally, we use the feature with the lowest entropy for object labeling.}
    \label{fig:system-overview}
\end{figure*} 

\noindent\textbf{The Effect of Prompt List on Entropy}. Following from the previous analysis, we study the effect of different prompt lists on entropy. We refer to this prompt list as the \textit{entropy prompt list}. Fig.~\ref{fig:mv_feature_fusion_improved} presents the object classification results when using different entropy prompt lists and the minimum entropy approach for feature selection. We do not include the weighted average as it was unaffected by prompt lists.

The results confirm that the prompt list influences entropy and can boost classification accuracy by up to 5\%. Using the ground-truth object list for each dataset achieves the highest accuracy; for example, the first 500 objects from the ScanNet++ dataset improve classification on ScanNet++ scenes by 5\% and the Replica list improves classification on the Replica dataset. Notably, the Replica list also generalizes, improving accuracy when used on ScanNet++ scenes. This suggests an optimal, balanced prompt list may exist which can be used for a variety of environments. We opted to use the Replica list as our entropy prompt list for all further experiments.

\section{Design of a Minimal System}
\label{sec:method}

In this section, we describe how our findings inform the design of a simple open-vocabulary system, focusing on the graph nodes. The system overview is presented in Fig.~\ref{fig:system-overview}.

\subsection{3D Mapping \& Segmentation}
Our system takes a stream of RGB-D images and poses as input to produce a 3D point cloud. This can be achieved by means of a 3D reconstruction or SLAM system. We then segment the output point cloud using the \emph{region growing} algorithm~\cite{Rusu_ICRA2011_PCL}, which groups points into regions based on proximity, surface normal consistency, or curvature. We run it with a tree search method to find 100 neighbors for each point, we set a smoothness threshold of \SI{0.05}{\radian} and a curvature threshold of 1.

Note that we preferred this geometry-based method because it provided a class-free segmentation output.  We did not evaluate closed-set 3D segmentation networks like Mask3D~\cite{schult2023mask3dmasktransformer3d} because the pre-trained labels tend to bias the outputs. Similarly, we did not use class-free 2D segmentation methods like SAM because of the significant computation overhead they introduced, as we discussed in Sec.~\ref{subsec:pre-processing}. However, newer class-free works such as SAM3D~\cite{yang2023sam3d} could provide additional benefits.

\subsection{Per-Segment View Association \& Scaling}

Each point in the point cloud contains its position, segment number, and the corresponding image from which it was collected. For each segment, we collate all associated images. The segment points are then reprojected onto the 2D image plane for each associated image, and a bounding box is drawn around the 2D points. This bounding box is scaled by a factor, and the image is cropped accordingly. By avoiding any pre-processing steps such as using multiple image scales or applying SAM masks, our approach is more suitable for real-time operation.

\begin{figure*}[ht]
    \centering
    \includegraphics[width=\textwidth]{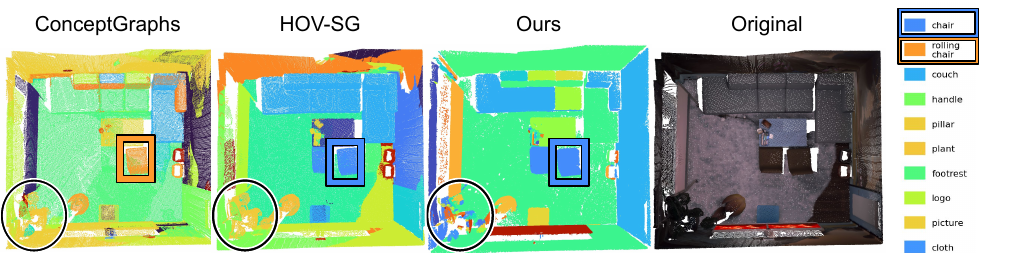}
    \caption{\small \textbf{Comparison of 3D semantic segmentation results for ConceptGraphs, HOV-SG and our method on Replica Office 0.} Our method tends to oversegment (lower left), while HOV-SG and ConceptGraphs undersegment. Differences in semantic labels (center rectangle) show that CLIP may struggle to distinguish between semantically similar concepts. The legend shows the top 10 labels in the scene. For more visualizations using the full label list or labels grouped into categories, see the Supp.~Section~\ref{sec:final_results}.}
    \label{fig:results-comparison-topdown}
\end{figure*} 

\subsection{Feature Selection}
Once the per-view features of each segment have been obtained, we compare the feature to a text encoding produced from the Replica object list using cosine similarity to obtain the \textit{inter-concept similarity} $s^{\text{concept}}_t$. This is a probability distribution across the object labels $i$. We calculate the entropy of this probability distribution as follows: 
\begin{equation}
\label{eq:entropy}
H = - \sum_{i=1}^{n} s^{\text{concept}}_t(i) \log(s^{\text{concept}}_t(i)),
\end{equation}
where $H$ is the entropy of the feature and $n$ is the number of prompt labels. As shown Sec.~\ref{subsec:feature-selection}, feature selection using a predefined prompt list works as a prior that improves performance by selecting the most informative view.

\section{Evaluation \& Discussion}
\label{sec:experiments}

In this section, we compare our proposed method to HOVSG and ConceptGraphs. Our goal is to achieve comparable performance while achieving a design that is more feasible for real-time applications.

\noindent\textbf{Experimental Setup.} We ran ConceptGraphs and our method on an AMD Ryzen Threadripper 1950X 16-Core Processor and NVIDIA GeForce RTX 2080 Ti. HOVSG was run on a larger desktop machine. The FPS figure reported for it in Tab.~\ref{tab:sota_comparison} is only an approximate indication of runtime rather than a direct comparison. For our evaluation, we used the Replica scenes \textit{office0-office4} and \textit{room0-room2}, using ground truth poses and depth images from the sequences. We did not use a SLAM or reconstruction method but instead accumulated the point clouds per scene using the ground truth poses. All methods use the ViT-H-14 CLIP backbone. ConceptGraphs uses MobileSAM~\cite{mobile_sam} and HOVSG uses the original SAM model.

To evaluate the methods, we followed the same methodology of our study (Sec.~\ref{sec:study}) by comparing the open-vocabulary features of each object against the extended list of 1687 labels from Replica and ScanNet++. The predicted labels for each object segment were evaluated against the Replica ground-truth labels to determine instance segmentation metrics such as the mIOU, F-mIOU, and mAcc. For more details, see Supp. Sec~\ref{sec:eval_metrics}. 

\noindent\textbf{Results.} Table~\ref{tab:sota_comparison} shows the results reported for the different methods. Our system produced comparable results to state-of-the-art (SoTA) methods when using optimal crop sizes for each scene (see Supp. Sec~\ref{sec:crop_sizes}). 
Our method requires a fraction of the computational cost of SoTA, demonstrating that even when using more efficient versions of SAM, SAM provides minimal benefit. For a breakdown of our computation times using a mid-range laptop see Supp. Sec.~\ref{sec:comp_times}.

Fig~\ref{fig:results-comparison-topdown} presents top-down views of the segmentation results from Replica \emph{Office 0} (additional examples are shown in Supp. Sec~\ref{sec:final_results}). We note that our method tends to over-segment objects when compared to HOVSG and ConceptGraphs. This is highlighted by the circled areas in Fig.~\ref{fig:results-comparison-topdown}. These differences can be explained by the segmentation methods, ours is purely geometry-based whereas the other methods merge segments across views. In particular, we observed that our approach will divide a sofa into its individual components, such as sofa cushions, whereas HOVSG may merge smaller items, like pillows, into a single segment. 

We can gain further insight into the low mIOU, F-mIOU, and mAcc scores by visually inspecting the semantic classes predicted by each method. CLIP tends to respond similarly across semantically related categories. For instance, as shown in the colored boxes in Fig.~\ref{fig:results-comparison-topdown}, labels like ``rolling chair'' and ``chair'' may be used interchangeably for the same object depending on the method, highlighting the difficulty of distinguishing closely related terms. Taking this into account, we show the segmentations grouped by broader categories in Supp Sec.~\ref{sec:final_results} for easier comparison.

\noindent\textbf{Discussion.} Our results could be improved by incorporating alternative class-free 3D segmentation methods, such as SAM3D. These methods offer potential solutions for segmenting complex shapes which are challenging to segment accurately using geometry alone, and could also help reduce the issue of over-segmentation. 

Our results also highlight challenges with evaluating CLIP. CLIP can frequently assign multiple, semantically similar labels to the same object, all of which may be valid. This flexible labeling aligns closely with human perception, where objects often have multiple names. For example, a sofa could equally be labeled as a couch, settee, or divan. While this variability demonstrates CLIP's capacity for generalization, it also introduces challenges in evaluation when using strict, closed-vocabulary metrics. These rigid evaluation metrics may obscure a model's true performance by penalizing labels that may be acceptable, which suggests the need for more flexible evaluation metrics. 

\begin{table}
\centering
\caption{\small \textbf{Comparison with SoTA.} We report average results across eight Replica scenes. We achieve comparable performance in 3D instance segmentation tasks for a fraction of the computational cost. For more details, see Supp. Sec.~\ref{sec:crop_sizes}.}
\label{tab:sota_comparison}
\footnotesize
\setlength{\tabcolsep}{1pt}
\begin{tabular}{L{1.1cm} C{0.8cm} C{1.1cm} C{0.8cm} C{1.1cm} C{1.6cm} C{1.2cm}}
\toprule
\multicolumn{1}{c}{\textbf{Method}} & \textbf{mIOU} & \textbf{F-mIOU} & \textbf{mAcc} & \textbf{FPS} & \textbf{Max VRAM} \\ \midrule
CG                                  & 0.06          & 0.10            & 0.08          & 0.16         & 8GB               \\
HOVSG                               & \textbf{0.08} & {\ul 0.13}      & \textbf{0.10} & 0.03         & 24GB              \\
Ours                                & {\ul 0.07}    & \textbf{0.14}   & {\ul 0.09}    & 0.51         & 5GB               \\ \bottomrule
\end{tabular}%
\end{table}

\section{Conclusion}
\label{sec:conclusion}

We presented a framework to study current open-vocabulary scene graphs and critical design choices. Across three studies, we found that image pre-processing strategies add significant computation time without improving classification, averaging features across views reduces accuracy, while feature selection offers an optimal balance---boosting performance without extra computational cost. We integrated our findings into a minimal pipeline that is able to achieve performance comparable to SoTA while achieving a threefold reduction in computational cost.

{
    \small
    \bibliographystyle{ieeenat_fullname}
    \bibliography{main}
}
\newpage
\clearpage
\setcounter{page}{1}

\maketitlesupplementary

\section{A Study of Open-Vocabulary Scene Graph Methods -- Further Details}
\label{sec:study-further-details}

In this section, we provide additional details about our study of open-vocabulary scene graph methods (Sec.~\ref{sec:study}). In particular, we list the specific ScanNet++ scenes used and describe the evaluation metrics in more detail. We provide a full results table for our image pre-processing study. Additionally, we present some examples which demonstrate the variance of CLIP responses which occur when the same object is seen from slightly different viewpoints.

\subsection{Methodology: Experimental Setup}
\label{sec:compute}
The studies (Sec.~\ref{subsec:pre-processing}, Sec.~\ref{subsec:feature-fusion}, and Sec.~\ref{subsec:feature-selection}) were all carried out on a mid-range laptop with an Intel i7 11850H @ 2.50GHz x 16, with an Nvidia RTX A3000 GPU. 

\subsection{Methodology: Datasets}
\label{sec:datasets}
Here, we provide more details about the scenes chosen from ScanNet++ for our studies. We chose 10 representative scenes outlined in Tab.~\ref{tab:scannetpp}. We used the provided iPhone RGB and depth images, as well as the ground truth segmented mesh of each scene. We processed the segmented mesh with the ScanNet++ Toolbox to obtain a \emph{ground truth segmented pointcloud}, used in our studies and experiments. 

\begin{table}[h]
\footnotesize
\centering
\caption{\small \textbf{Full list of chosen ScanNet++ scenes and scene types}. The scenes are three student bedrooms, two offices, two apartments, one lab, one classroom, and one student common room.}
\label{tab:scannetpp}
\begin{tabular}{@{}ll@{}}
\toprule
\textbf{Scene ID} & \textbf{Scene Type} \\ \midrule
\textit{0a7cc12c03} & Student Bedroom \\
\textit{0a76e06478} & Student Bedroom \\
\textit{1d003b07bd} & Lab \\
\textit{2b1dc6d6a6} & Office \\
\textit{8a35ef3cfe} & Apartment \\
\textit{fd361ab85f} & Office \\
\textit{ef69d58016} & Apartment \\
\textit{c173f62b15} & Student Common Room \\
\textit{5a269ba6fe} & Classroom \\
\textit{7e09430da7} & Student Bedroom \\ \bottomrule
\end{tabular}
\end{table}

\subsection{Methodology: Ground-truth Crop Generation}
\label{sec:crop_generation}
To study the effect of crops and masks in the pre-processing step, we produced a set of \emph{ground truth image crops} for each object in the scene. To achieve this, we iterated over each object instance in the ground truth point cloud and projected their 3D points onto every image of the scene where the instance was visible. We used the projected 2D points on each image to determine a bounding box enclosing the object. The bounding box was scaled and cropped using the factors discussed in Sec.~\ref{sec:study}: 1.0, 1.2, 1.5, 1.8, and 2.0, effectively generating a set of \emph{ground truth multi-scale crops} with different amounts of context.

We also produced \emph{ground truth object masks} to evaluate the performance of masking the objects using SAM2. We input SAM2 with the same projected 2D points (from the bounding boxes) as \emph{click prompts} to segment the object of interest. The segments we produced were generated with different background colors as discussed in prior work---white, black, and transparent (alpha = 0).

\subsection{Study 1: Per-Object Classification Accuracy}
\label{sec:classification_accuracy}

We first define per-object classification accuracy. Let \( N \) be the total number of objects and \( V_i \) be the number of views for object \( i \), where \( i = 1, 2, \dots, N \). An indicator variable \( c_{ij} \) is used to indicate if the classification for view \( j \) of object \( i \) is correct (1) or incorrect (0).

The classification accuracy for an individual object \( i \) is a percentage given as:
$$
\text{Accuracy}_i = \frac{1}{V_i} \sum_{j=1}^{V_i} c_{ij}
$$
The overall classification accuracy across all objects is:
\[
\text{Overall Accuracy} = \frac{1}{N} \sum_{i=1}^{N} \left( \frac{1}{V_i} \sum_{j=1}^{V_i} c_{ij} \right)
\]
\subsection{Study 1: Image Pre-processing Results Tables}
\label{sec:preproc_table}

We provide complete result tables for our image-processing study (Sec.~\ref{subsec:pre-processing}). Tab.~\ref{tab:crops} presents the impact of scaling crop sizes, Tab.~\ref{tab:multi-scale} shows the effect of fusing multi-scale crops per view, and Tab.~\ref{tab:sam} illustrates the influence of SAM masks and the combined effect of SAM and crop fusion on object classification accuracy. We also present the computation time for each of these methods in Tab.~\ref{tab:computation_time}. 

Of the considered image pre-processing methods, the most effective approach combines a SAM2 mask on a black background with an image crop scaled to 1.5x the original bounding box and achieves an accuracy of 14.6\% (Tab.~\ref{tab:sam}). Notably, using only SAM2 masks or only scaled crops achieves similar accuracy: 12.2\% for SAM2 masks with a black background, and 11.3\% for 1.5x scaled crops, see Tab.~\ref{tab:sam} and Tab.~\ref{tab:crops} respectively. 

Regarding computation time, we identified significant differences between cropping and masking, shown in Tab.~\ref{tab:computation_time}. In particular, single-scale cropping requires an average time of 78ms for cropping and CLIP extraction per object viewpoint, with most of the time spent on CLIP. Conversely, SAM and CLIP extraction requires 223ms on average. These numbers scale linearly with the number of combinations considered.

As a result of these findings, we opt to use only scaled crops, reducing computation time to approximately \SI{78}{\milli\second} per viewpoint while achieving a classification accuracy of 11.3\%. This choice significantly improves processing speed with only a minor reduction in accuracy. Further, we note that in an online setting, processing more views (i.e., integrating more data) could be more useful for producing optimal performance than \emph{augmenting} the existing ones.

\begin{table}[ht]
\centering
\caption{\small{\textbf{Computation time for the evaluated pre-processing approaches (cropping and masking)}. The computation times are averaged over 100 runs. The time reported for cropping mostly reflects the CLIP forward pass; the time for masking includes SAM and CLIP extraction. Combinations of different scales, or scaling and masking are indicated with the + symbol, these scaled additively with the base times.}}
\label{tab:computation_time}
\resizebox{\columnwidth}{!}{%
\begin{tabular}{@{}ccc@{}}
\toprule
\textbf{Type} & \textbf{Combinations} & \begin{tabular}[c]{@{}c@{}}\textbf{Computation}\\ \textbf{Time (ms)}\end{tabular} \\ \midrule
\multirow{5}{*}{Crops} & 1.0 & 73.5 \\
 & 1.2 & 79.5 \\
 & 1.5 & 78.1 \\
 & 1.8 & 77.9 \\
 & 2.0 & 77.8 \\ \midrule
\multirow{3}{*}{Multi-Scale Crop Fusion} & 1.0 + 1.5 & 165 \\
 & 1.2 + 1.5 + 1.8 & 220 \\
 & 1.0 + 1.2 + 1.5 + 1.8 & 287 \\ \midrule
\multirow{3}{*}{SAM Masks} & White & 223 \\
 & Black & 223 \\
 & Transparent & 223 \\ \midrule
\multirow{4}{*}{SAM and Crop Fusion} & Black + 1.5 & 314 \\
 & Black + 1.0 + 1.5 & 369 \\
 & Black + 1.2 + 1.5 + 1.8 & 436 \\
 & Black + 1.0 + 1.2+ 1.5 + 1.8 & 505 \\ \bottomrule
\end{tabular}%
}
\end{table}

\subsection{Study 2: Multi-View Variance of CLIP: Further Examples}
\label{sec:multi-view-variance-examples}

\begin{figure}[h]
    \centering
    \includegraphics[width=1\columnwidth, clip, trim={0 0 0 0}]{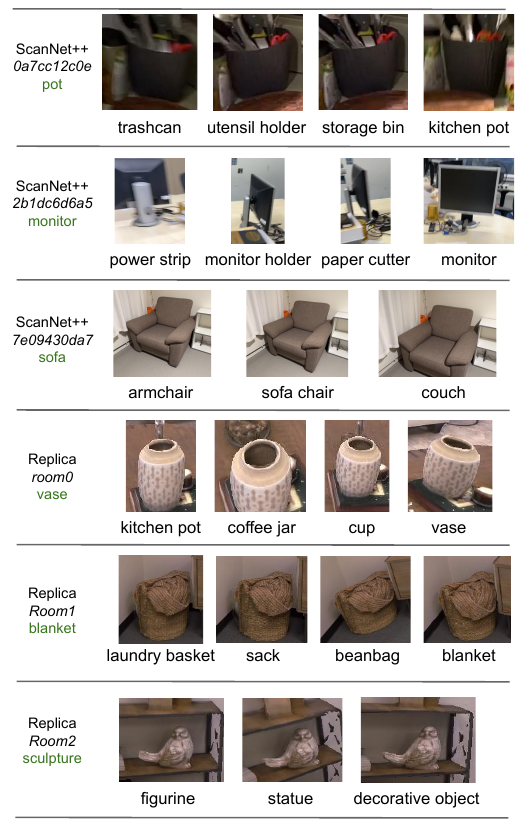}
    \caption{\small \textbf{Examples illustrating multi-view variance in CLIP outputs.} The green label on the left is the ground truth object label. Despite minimal changes in viewpoint, CLIP responses can still exhibit significant variation.} 
    \label{fig:mv_clip}
\end{figure}

In Fig.~\ref{fig:mv_clip}, we show additional examples of the per-object multi-view variability of CLIP outputs discussed in Sec.~\ref{subsec:feature-fusion} (Fig.~\ref{fig:mv_variance}). We present three examples from different ScanNet++ scenes and three from Replica scenes. These examples illustrate how CLIP's predictions can vary significantly across different viewpoints of an object and even among very similar viewpoints. The variation can range from synonyms, as seen in ScanNet++ scene \textit{7809}, where similar views of the same object are labeled as ``armchair'', ``sofa chair'' and ``couch'', to more substantial differences, as in ScanNet++ scene \text{2b1d}, where different views of a monitor are labeled as ``power strip'' and ``paper cutter''.

In many cases, only one or a few specific viewpoints provide the correct label for an object, suggesting that averaging features across the views may not be the optimal approach for fusing information. This motivated the feature selection approach we proposed in Study~\ref{subsec:feature-selection} which prioritizes the feature and viewpoint that \emph{best} represents the object in question.

In the case of ScanNet++ scene \textit{2b1d} and Replica scene \textit{Room 2} in particular, the labels vary but are semantically similar. For example, ``armchair'', ``chair'' and ``couch'' could be considered to be synonyms of each other. We discuss this further both in Sec.~\ref{sec:experiments} and in Sec.~\ref{sec:final_results}.

\section{SoTA Comparison Experiments -- Further Details}

In this section, we present additional details on the performance of our method when compared to ConceptGraphs and HOVSG, described in Sec.~\ref{sec:experiments} (Tab.~\ref{tab:sota_comparison}). We begin by defining the evaluation metrics used in our study, followed by a full results table that illustrates the impact of crop size on our final segmentation accuracy (Tab.~\ref{tab:full-results}). We also provide qualitative point cloud segmentations of the Replica scenes for all methods to further illustrate our findings.

\subsection{3D Semantic Segmentation Evaluation Metrics}
\label{sec:eval_metrics}

For our comparison with the state-of-the-art, we use standard 3D semantic segmentation metrics of mIOU, F-mIOU, and mAcc, defined as follows:

\begin{equation}
    \text{mIOU} = \frac{1}{N} \sum_{i=1}^{N} \frac{T_{P_i}}{T_{P_i} + F_{P_i} + F_{N_i}} \tag{1},
\end{equation}

\begin{equation}
    \text{F-IOU} = \frac{1}{\sum_{N} n_i} \sum_{i=1}^{N} \frac{n_i \cdot T_{P_i}}{T_{P_i} + F_{P_i} + F_{N_i}} \tag{2},
\end{equation}

\begin{equation}
    \text{mAcc} = \frac{1}{N} \sum_{i=1}^{N} \frac{T_{P_i}}{T_{P_i} + F_{P_i}} \tag{3},
\end{equation}

\subsection{3D Semantic Segmentation Results Tables}
\label{sec:crop_sizes}
Tab.~\ref{tab:full-results} shows how our results vary with different crop sizes, complementing the results in Tab.~\ref{tab:sota_comparison}. Due to the variety of object sizes and shapes in each scene and the tendency of region growing to over-segment objects, the optimal crop size differs by scene. Therefore, using a single crop size across all scenes reduces performance. We argue that a better approach might be to automatically adjust crop size based on an object's distance from the camera and its size, allowing greater flexibility to handle imperfect segments.
\begin{table}[h]
\centering
\footnotesize
\caption{\small \textbf{Impact of crop size on the 3D instance segmentation performance achieved by our method for Replica.} By fine-tuning the crop scale by scene we can maximize performance.}
\label{tab:full-results}
\begin{tabular}{@{}cccc@{}}
\toprule
\textbf{Crop Size} & \textbf{mIOU} & \textbf{F-mIOU} & \textbf{mAcc} \\ \midrule
1 & {\ul \textit{0.041}} & {\ul \textit{0.093}} & {\ul \textit{0.064}} \\
1.2 & 0.038 & 0.075 & 0.054 \\
1.5 & 0.039 & 0.063 & 0.055 \\
1.8 & {\ul \textit{0.050}} & {\ul \textit{0.076}} & {\ul \textit{0.066}} \\
2 & 0.033 & 0.046 & 0.044 \\
\begin{tabular}[c]{@{}c@{}} Optimal Size\end{tabular} & \textbf{0.065} & \textbf{0.138} & \textbf{0.089} \\ \bottomrule
\end{tabular}%
\end{table}

\subsection{Computation Time Breakdown}
\label{sec:comp_times}
In this section, we present a breakdown of the computation time required by our method. Our goal is to show that even on more minimal hardware we can achieve close to real-time performance. We measured the computation time on a mid-range laptop with an Intel i7 11850H @ 2.50GHz x 16 with an NVIDIA RTX A3000 GPU. Tab.~\ref{tab:comp_times} shows the results for eight Replica scenes.

As anticipated, region growing requires a fixed time budget for scenes with a similar number of points. The majority of the computation---about 85\%---is allocated to feature extraction and selection. The required computation time is influenced by factors such as the number of frames sampled from the trajectory, the number of views retained per object, and the size of the point cloud. In our setup, we sample every 50 frames to construct the point cloud, retain all views per object, and downsample the point cloud before region growing. Computation could be further reduced by additional downsampling or by limiting the number of views per object, though this may come at the cost of object classification performance.

\begin{table}[ht]
\centering
\caption{\small \textbf{Computation time required by our method on Replica scenes.} By removing requirements for 2D segmentation models such as SAM and selecting multi-view features
as opposed to fusing, we can greatly reduce computation.}
\label{tab:comp_times}
\resizebox{\columnwidth}{!}{%
\begin{tabular}{@{}lccccl@{}}
\toprule
 & \begin{tabular}[c]{@{}c@{}}\textbf{Point Cloud}\\ \textbf{Size}\end{tabular} & \begin{tabular}[c]{@{}c@{}}\textbf{Region} \\ \textbf{Growing} \\ \textbf{[s]}\end{tabular} & \begin{tabular}[c]{@{}c@{}}\textbf{Feature Extraction}\\ \textbf{\&} \textbf{Selection}\\ \textbf{[s]}\end{tabular} & \begin{tabular}[c]{@{}c@{}}\textbf{Total Time} \\ \textbf{[s]}\end{tabular} & \textbf{FPS} \\ \midrule
\textit{room0} & 927,067 & 28.9 & 211 & 240 & 0.17 \\
\textit{room1} & 979,194 & 20.2 & 94.7 & 115 & 0.35 \\
\textit{room2} & 979,155 & 22.1 & 123 & 145 & 0.28 \\
\textit{office0} & 978,719 & 19.4 & 111 & 130 & 0.31 \\
\textit{office1} & 979,188 & 22.6 & 164 & 187 & 0.21 \\
\textit{office2} & 964,842 & 20.5 & 116 & 137 & 0.29 \\
\textit{office3} & 964,334 & 24.7 & 157 & 182 & 0.22 \\
\textit{office4} & 979,191 & 18.5 & 97.8 & 116 & 0.34 \\ \midrule
Mean & 968,961 & 22.1 & 134 & 156 & 0.27 \\ \bottomrule
\end{tabular}%
}
\end{table}

\subsection{3D Semantic Segmentation Further Results}
\label{sec:final_results}
In the following pages, we present top-down segmentation examples for a selection of Replica scenes we tested. We colored the point clouds using two methods:

\begin{enumerate}
\item \textbf{Full Evaluation Prompt List.} At the top of each figure, we show the point clouds colored using predicted labels from the full evaluation prompt list of 1687 labels. Below each scene, we list the 30 most common labels. We highlight two key aspects of the results: variations in segmentation (circles) and differences in predicted labels (rectangles). 
\item \textbf{Grouped Labels.} At the bottom of each figure, we group the labels into broader categories such as ``Kitchen \& Dining'' and ``Seating'' to make it easier to compare the three methods. We based these categories on shopping categories from websites such as Amazon.  We first collate the unique labels chosen by all methods for all Replica scenes (approximately 300 labels) and then manually assign objects to these categories. The categories and groupings are shown in Tab.~\ref{tab:categories}.
\end{enumerate}

\noindent\textbf{Discussion.} For the segmentation module, our region growing component tends to over-segment. By comparison, HOVSG and ConceptGraphs under-segment as they merge segments over time. For example, in \textit{room 2}, our method segments each chair into individual components, while ConceptGraphs and HOVSG group each chair as a single segment. In \textit{office 1}, our method segments the pillows into smaller segments, whereas ConceptGraphs and HOVSG merge them into one. Improvements to our segmentation module are possible and are intended as future work.

The labels predicted by each method, while initially appearing quite different, are often semantically similar on closer inspection. This is especially apparent when looking at the grouped category segmentation. We illustrate these differences with colored boxes. For example, in \textit{room 1}, all methods identify related labels for the cabinet, such as ``small cabinet'', ``shelf'', and ``drawer''. In \textit{office 1}, the TV is labeled as both ``monitor'' and ``picture'', and in \textit{office 3}, the armchair is identified as both ``easy chair'' and ``seat''. This likely explains the low mIOU, F-mIOU, and mAcc results across all three methods and why open-vocabulary methods, in general, seem to underperform when compared to fixed-class models trained specifically for object classification.

Easier comparisons between the methods can be made when inspecting the grouped segmentation. For example, in \textit{room 1}, the labels chosen in the top figure appear to be different, but the bottom figure shows that the bed and pillows are in the same ``Bedroom'' and ``Pillows \& Cushions'' category across all methods. Similarly, the cabinet labels across methods are in the ``Storage \& Shelves'' category. 

The semantically similar labels selected by these methods also highlight our limited understanding of both the limitations and potential of open-vocabulary models for 3D object classification. Does CLIP exhibit biases toward specific object labels, such as favoring ``sofa'' over ``couch'' or vice versa? Could the true performance of these systems be obscured by restricting the range of prompts or limiting our ground truth labels? Developing more flexible and nuanced evaluation frameworks that capture the full versatility of language may be necessary to address these questions effectively.

\begin{table*}[ht]
\centering
\caption{\small \textbf{Study 1: Effect of crop size on object classification accuracy (\%) in scenes from ScanNet++ and Replica.} Crop sizes scaled to 1.2x and 1.5x of the object bounding box achieve the best accuracy (using ground truth segments), while larger sizes reduce performance due to irrelevant contextual information.}
\label{tab:crops}
\resizebox{\textwidth}{!}{%
\begin{tabular}{@{}cccccccccccccccccccc@{}}
\toprule
\multirow{2}{*}{\textbf{Crop Sizes}} & \multicolumn{10}{c}{\textbf{ScanNet++ Scenes}} & \multicolumn{8}{c}{\textbf{Replica Scenes}} & \multicolumn{1}{l}{\multirow{2}{*}{\textbf{Mean}}} \\
 & \multicolumn{1}{l}{\textit{0a7c}} & \multicolumn{1}{l}{\textit{0a76}} & \multicolumn{1}{l}{\textit{1d003}} & \multicolumn{1}{l}{\textit{2b1d}} & \multicolumn{1}{l}{\textit{8a35}} & \multicolumn{1}{l}{\textit{fd36}} & \multicolumn{1}{l}{\textit{ef69}} & \multicolumn{1}{l}{\textit{c173}} & \multicolumn{1}{l}{\textit{5a26}} & \multicolumn{1}{l}{\textit{7e09}} & \multicolumn{1}{l}{\textit{room0}} & \multicolumn{1}{l}{\textit{room1}} & \multicolumn{1}{l}{\textit{room2}} & \multicolumn{1}{l}{\textit{office0}} & \multicolumn{1}{l}{\textit{office1}} & \multicolumn{1}{l}{\textit{office2}} & \multicolumn{1}{l}{\textit{office3}} & \multicolumn{1}{l}{\textit{office4}} & \multicolumn{1}{l}{} \\ \midrule
1 & 15.7 & 14.3 & 14.5 & 8.4 & 17.6 & 11.6 & 7.6 & 9.5 & 9.1 & 4.6 & 22.5 & 12.6 & 15.2 & 7.9 & 4.8 & 8.9 & 5.8 & 3.4 & 10.3 \\
\rowcolor{TableGray}
1.2 & 16.5 & 14.6 & 17.3 & 8.9 & 18.0 & 13.7 & 10.6 & 10.2 & 9.6 & 8.0 & 24.2 & 12.1 & 19.5 & 7.1 & 3.3 & 8.9 & 6.0 & 4.9 & \textbf{11.3} \\
\rowcolor{TableGray}
1.5 & 17.5 & 13.7 & 22.9 & 8.4 & 19.3 & 15.7 & 10.4 & 8.6 & 7.4 & 7.3 & 24.5 & 12.3 & 19.6 & 4.7 & 2.2 & 7.5 & 5.7 & 5.3 & \textbf{11.3} \\
1.8 & 17.4 & 12.1 & 26.6 & 7.4 & 18.9 & 13.4 & 9.9 & 7.7 & 4.9 & 7.6 & 20.3 & 11.7 & 18.3 & 4.8 & 0.9 & 6.8 & 4.4 & 5.9 & 10.6 \\
2 & 15.7 & 10.9 & 21.4 & 7.4 & 17.4 & 12.6 & 8.7 & 9.9 & 3.3 & 6.6 & 15.7 & 11.3 & 15.2 & 4.1 & 0.9 & 5.8 & 4.4 & 4.6 & 9.3 \\ \bottomrule
\end{tabular}%
}
\end{table*}

\begin{table*}[ht]
\centering
\caption{\small \textbf{Study 1: Impact of multi-scale crop fusion on object classification accuracy (\%) for Replica and ScanNet++.} Using three scales achieves the highest accuracy with no further benefit seen with additional scales. In Tab.~\ref{tab:computation_time}, we additionally show that the computation time grows proportionally with the number of scales considered.}
\label{tab:multi-scale}
\resizebox{\textwidth}{!}{%
\begin{tabular}{@{}crrrrrrrrrrrrrrrrrrc@{}}
\toprule
\multirow{2}{*}{\textbf{\begin{tabular}[c]{@{}c@{}}Multi-Scale \\ Crop Combinations\end{tabular}}} & \multicolumn{10}{c}{\textbf{ScanNet++ Scenes}} & \multicolumn{8}{c}{\textbf{Replica Scenes}} & \multicolumn{1}{l}{\multirow{2}{*}{\textbf{Mean}}} \\
 & \multicolumn{1}{l}{\textit{0a7c}} & \multicolumn{1}{l}{\textit{0a76}} & \multicolumn{1}{l}{\textit{1d003}} & \multicolumn{1}{l}{\textit{2b1d}} & \multicolumn{1}{l}{\textit{8a35}} & \multicolumn{1}{l}{\textit{fd36}} & \multicolumn{1}{l}{\textit{ef69}} & \multicolumn{1}{l}{\textit{c173}} & \multicolumn{1}{l}{\textit{5a26}} & \multicolumn{1}{l}{\textit{7e09}} & \multicolumn{1}{l}{\textit{room0}} & \multicolumn{1}{l}{\textit{room1}} & \multicolumn{1}{l}{\textit{room2}} & \multicolumn{1}{l}{\textit{office0}} & \multicolumn{1}{l}{\textit{office1}} & \multicolumn{1}{l}{\textit{office2}} & \multicolumn{1}{l}{\textit{office3}} & \multicolumn{1}{l}{\textit{office4}} & \multicolumn{1}{l}{} \\ \midrule
1 + 1.2 & 17.9 & 15.7 & 20.6 & 9.1 & 18.3 & 13.3 & 9.3 & 9.7 & 10.4 & 5.9 & 24.2 & 11.9 & 16.4 & 7.7 & 4.6 & 9.3 & 6.3 & 3.5 & 11.9 \\
1 + 1.5 & 19.2 & 15.7 & 21.2 & 10.5 & 19.7 & 13.5 & 10.0 & 10.9 & 8.7 & 7.7 & 26.2 & 12.6 & 18.2 & 6.1 & 2.2 & 9.5 & 6.4 & 3.8 & 12.3 \\
1 + 1.8 & 20.0 & 14.3 & 24.6 & 9.8 & 20.3 & 18.2 & 10.4 & 10.6 & 6.7 & 8.3 & 26.4 & 13.0 & 20.0 & 5.5 & 1.7 & 9.6 & 6.7 & 3.5 & 12.8 \\
1 + 2 & 18.6 & 14.7 & 21.4 & 8.6 & 19.9 & 18.4 & 10.4 & 11.1 & 4.5 & 8.6 & 25.6 & 13.3 & 19.3 & 6.1 & 0.9 & 10.4 & 6.8 & 3.5 & 12.3 \\
1.2 + 1.5 & 17.7 & 14.2 & 21.8 & 8.9 & 19.4 & 15.8 & 11.3 & 10.1 & 8.4 & 7.6 & 26.1 & 12.1 & 19.4 & 5.9 & 3.3 & 8.6 & 5.8 & 5.7 & 12.3 \\
1.2 + 1.8 & 19.1 & 13.6 & 23.5 & 8.0 & 19.3 & 17.3 & 10.6 & 9.6 & 7.1 & 8.3 & 25.7 & 13.0 & 20.6 & 6.1 & 2.0 & 9.1 & 5.8 & 4.6 & 12.4 \\
1.2 + 2 & 18.2 & 12.9 & 21.6 & 8.4 & 19.1 & 18.0 & 10.8 & 10.1 & 5.6 & 8.8 & 23.5 & 13.0 & 21.2 & 5.8 & 0.9 & 8.7 & 5.4 & 5.0 & 12.1 \\
1.5 + 1.8 & 18.6 & 12.9 & 27.5 & 8.3 & 19.5 & 16.2 & 10.2 & 8.3 & 6.6 & 7.8 & 24.7 & 12.6 & 20.2 & 4.5 & 0.9 & 7.9 & 4.6 & 5.7 & 12.1 \\
1.5 + 2.0 & 18.5 & 12.4 & 24.5 & 8.2 & 19.0 & 16.6 & 9.6 & 8.5 & 5.4 & 7.4 & 23.2 & 12.4 & 19.6 & 4.5 & 0.2 & 7.6 & 4.9 & 4.1 & 11.5 \\
1.8 + 2.0 & 16.3 & 11.7 & 21.0 & 7.5 & 18.5 & 12.9 & 9.2 & 8.5 & 4.3 & 7.1 & 19.5 & 11.6 & 18.1 & 4.4 & 0.0 & 6.7 & 4.4 & 4.7 & 10.4 \\
1 + 1.2 + 1.5 & 18.0 & 15.3 & 21.5 & 9.8 & 18.7 & 14.4 & 10.9 & 10.3 & 9.2 & 7.8 & 25.9 & 12.6 & 18.8 & 6.4 & 3.7 & 9.7 & 6.5 & 4.0 & 12.4 \\
\rowcolor{TableGray}
1+ 1.2 + 1.8 & 19.7 & 14.5 & 22.9 & 9.8 & 19.7 & 17.3 & 10.8 & 10.6 & 8.1 & 8.6 & 26.8 & 13.6 & 20.2 & 6.5 & 2.2 & 10.1 & 6.9 & 3.8 & \textbf{12.9} \\
1 + 1.2 + 2.0 & 19.4 & 15.4 & 19.5 & 8.9 & 18.8 & 18.7 & 11.0 & 11.0 & 6.7 & 9.3 & 26.5 & 13.4 & 20.4 & 6.5 & 1.5 & 9.7 & 6.9 & 3.8 & 12.6 \\
1 + 1.5 + 1.8 & 20.2 & 14.4 & 22.2 & 8.9 & 20.4 & 17.4 & 10.8 & 10.9 & 6.7 & 8.2 & 27.4 & 12.9 & 19.3 & 5.5 & 1.7 & 9.8 & 6.6 & 4.6 & 12.7 \\
1 + 1.5 + 2.0 & 20.0 & 15.0 & 26.3 & 9.2 & 20.0 & 18.2 & 10.6 & 11.1 & 6.4 & 8.4 & 26.9 & 13.1 & 19.6 & 5.0 & 1.7 & 9.0 & 6.9 & 4.4 & \textbf{12.9} \\
1 + 1.8 + 2.0 & 18.0 & 14.9 & 25.0 & 8.2 & 20.3 & 17.7 & 10.2 & 9.9 & 5.4 & 8.5 & 26.2 & 13.1 & 19.9 & 5.2 & 1.5 & 9.5 & 6.9 & 4.1 & 12.5 \\
1.2 + 1.5 + 1.8 & 19.3 & 14.1 & 24.0 & 8.3 & 19.5 & 17.4 & 10.9 & 9.1 & 7.3 & 7.6 & 26.1 & 12.4 & 20.2 & 5.2 & 2.0 & 8.6 & 5.7 & 5.1 & 12.4 \\
1.2 + 1.5 + 2.0 & 19.0 & 13.8 & 19.9 & 8.3 & 20.0 & 16.8 & 10.9 & 9.4 & 7.4 & 7.9 & 25.4 & 12.3 & 21.1 & 5.5 & 1.1 & 8.5 & 5.7 & 5.0 & 12.1 \\
1.2 + 1.8 + 2.0 & 18.0 & 13.2 & 20.4 & 8.3 & 19.7 & 15.3 & 10.9 & 9.3 & 5.5 & 8.0 & 24.4 & 12.7 & 20.7 & 5.2 & 1.3 & 8.4 & 5.0 & 4.9 & 11.7 \\
1.5 + 1.8 + 2.0 & 17.6 & 12.7 & 27.7 & 7.8 & 19.8 & 13.3 & 9.7 & 8.7 & 5.8 & 7.8 & 23.5 & 12.3 & 19.5 & 4.5 & 0.2 & 7.3 & 4.9 & 5.0 & 11.6 \\
\rowcolor{TableGray}
1 + 1.2 + 1.5 + 1.8 & 19.3 & 14.9 & 23.5 & 9.1 & 19.8 & 17.8 & 11.2 & 10.9 & 7.8 & 8.3 & 27.3 & 12.7 & 19.9 & 6.5 & 2.4 & 9.7 & 6.7 & 4.4 & \textbf{12.9} \\
1 + 1.2 + 1.5 + 2.0 & 19.0 & 15.1 & 24.4 & 9.3 & 18.8 & 17.5 & 11.2 & 10.7 & 7.3 & 8.6 & 27.1 & 13.0 & 20.1 & 6.2 & 1.7 & 9.9 & 6.9 & 4.3 & 12.8 \\
\rowcolor{TableGray}
1 + 1.2 + 1.8 + 2.0 & 19.1 & 14.4 & 25.2 & 9.0 & 20.5 & 18.6 & 10.7 & 11.3 & 6.5 & 8.7 & 26.3 & 13.1 & 21.1 & 5.8 & 1.5 & 10.0 & 6.7 & 4.3 & \textbf{12.9} \\
1 + 1.5 + 1.8 + 2.0 & 19.3 & 14.6 & 25.0 & 8.3 & 20.6 & 17.8 & 10.1 & 9.4 & 6.2 & 7.9 & 26.8 & 12.6 & 20.0 & 4.8 & 1.5 & 9.5 & 6.4 & 4.7 & 12.5 \\
1.2 + 1.5 + 1.8 + 2.0 & 18.0 & 13.6 & 25.5 & 8.4 & 19.8 & 16.3 & 10.5 & 8.9 & 6.7 & 7.9 & 25.2 & 12.7 & 20.6 & 4.8 & 0.9 & 8.3 & 5.3 & 5.0 & 12.1 \\
\rowcolor{TableGray}
1 + 1.2 + 1.5 + 1.8 + 2.0 & 19.4 & 14.4 & 25.1 & 8.8 & 20.3 & 17.6 & 10.9 & 10.9 & 7.1 & 9.5 & 27.1 & 12.7 & 20.0 & 5.8 & 1.3 & 9.9 & 6.6 & 4.6 & \textbf{12.9} \\ \bottomrule
\end{tabular}%
}
\end{table*}

\begin{table*}[t!]
\centering
\caption{\small \textbf{Study 1: Impact of SAM masks and SAM and crop fusion on object classification accuracy (\%) for Replica and ScanNet++.} A SAM mask with a black background when fused with a crop scaled to 1.5x the original bounding box offers the most improvement.}
\label{tab:sam}
\resizebox{\textwidth}{!}{%
\begin{tabular}{@{}cccccccccccccccccccc@{}}
\toprule
\multirow{2}{*}{\textbf{\begin{tabular}[c]{@{}c@{}}SAM Masks \& \\ SAM and Crop Fusion\end{tabular}}} & \multicolumn{10}{c}{\textbf{ScanNet++ Scenes}} & \multicolumn{8}{c}{\textbf{Replica Scenes}} & \multicolumn{1}{l}{\multirow{2}{*}{\textbf{Mean}}} \\
 & \multicolumn{1}{l}{\textit{0a7c}} & \multicolumn{1}{l}{\textit{0a76}} & \multicolumn{1}{l}{\textit{1d003}} & \multicolumn{1}{l}{\textit{2b1d}} & \multicolumn{1}{l}{\textit{8a35}} & \multicolumn{1}{l}{\textit{fd36}} & \multicolumn{1}{l}{\textit{ef69}} & \multicolumn{1}{l}{\textit{c173}} & \multicolumn{1}{l}{\textit{5a26}} & \multicolumn{1}{l}{\textit{7e09}} & \multicolumn{1}{l}{\textit{room0}} & \multicolumn{1}{l}{\textit{room1}} & \multicolumn{1}{l}{\textit{room2}} & \multicolumn{1}{l}{\textit{office0}} & \multicolumn{1}{l}{\textit{office1}} & \multicolumn{1}{l}{\textit{office2}} & \multicolumn{1}{l}{\textit{office3}} & \multicolumn{1}{l}{\textit{office4}} & \multicolumn{1}{l}{} \\ \midrule
SAM White & 15.2 & 12.1 & 18.8 & 9.6 & 18.4 & 16.1 & 9.6 & 10.5 & 7 & 6.5 & 20.6 & 15.1 & 14.8 & 2.7 & 4.8 & 7.1 & 4.8 & 5.0 & 11 \\
SAM Black & 17.3 & 11.5 & 19.3 & 10.6 & 19.4 & 16.8 & 9.8 & 12.7 & 9.5 & 7.6 & 16.5 & 17.4 & 17.5 & 5.8 & 8.9 & 5.2 & 5.4 & 8.9 & 12.2 \\
SAM Transparent & 18.2 & 13.9 & 16.7 & 10.2 & 19 & 14.9 & 8.5 & 12 & 8.2 & 8.5 & 15.8 & 16.4 & 13.9 & 4.4 & 7.3 & 4.2 & 4.5 & 5.9 & 11.3 \\
\rowcolor{TableGray}
\begin{tabular}[c]{@{}c@{}}SAM Black\\ 1.5\end{tabular} & 21.8 & 15.5 & 25.2 & 12.1 & 23.8 & 17.2 & 13.0 & 13.1 & 13.6 & 10.0 & 21.3 & 17.9 & 22.4 & 8.6 & 7.3 & 7.5 & 5.1 & 6.5 & \textbf{14.6} \\
\begin{tabular}[c]{@{}c@{}}SAM Black \\ 1 + 1.8\end{tabular} & 24.4 & 16.0 & 24.4 & 12.5 & 23.0 & 18.2 & 12.9 & 12.8 & 13.0 & 9.0 & 22.8 & 17.7 & 22.4 & 8.4 & 3.2 & 9.2 & 6.4 & 5.5 & 14.5 \\
\begin{tabular}[c]{@{}c@{}}SAM Black\\ 1.0 + 1.5 + 2.0\end{tabular} & 23.0 & 15.7 & 28.5 & 11.8 & 23.2 & 17.1 & 12.4 & 13.2 & 10.4 & 9.3 & 24.0 & 16.3 & 22.2 & 7.5 & 3.4 & 10.1 & 7.1 & 5.5 & 14.5 \\
\begin{tabular}[c]{@{}c@{}}SAM Black \\ 1 + 1.2 + 1.8 + 2.0\end{tabular} & 22.5 & 15.7 & 25.9 & 11.2 & 22.2 & 17.0 & 12.0 & 12.0 & 9.2 & 9.5 & 25.2 & 16.1 & 22.0 & 6.7 & 2.7 & 10.3 & 7.6 & 5.0 & 14.0 \\
\begin{tabular}[c]{@{}c@{}}SAM Black \\ 1 + 1.2 + 1.5 + 1.8 + 2.0\end{tabular} & 22.0 & 15.5 & 25.8 & 11.1 & 22.4 & 17.3 & 12.1 & 11.8 & 9.1 & 10.2 & 25.9 & 14.7 & 22.6 & 6.3 & 2.7 & 10.4 & 7.2 & 5.2 & 14.0 \\ \bottomrule
\end{tabular}%
}
\end{table*}

\begin{table*}
\caption{\textbf{Evaluation prompt list grouped into broader item categories}. We chose 17 categories that encompass the unique categories selected by all three methods (ours, ConceptGraphs, and HOVSG) on the Replica scenes. We collate any ceiling, glass, wall, and panel labels into background and any less common objects such as metal sheet into the miscellaneuous category.}
\resizebox{\textwidth}{!}{%
\begin{tabular}{@{}ll@{}}
\toprule
\textbf{Category} & \textbf{Labels} \\ \midrule
Kitchen \& Dining & \begin{tabular}[c]{@{}l@{}}beverage can, blender, bread toaster, canister, circular tray, cup, cutboard, dish, dispenser bottle, jug, \\ juice box, juice tetrapack, knife, jar, bottle, banana, cabbage, bin, egg carton, food container, garbage bin cover, \\ lid, pot, recycle bin, saucer, sink, soap, soap container, spoon, tissue, tissue box, tissue paper, tissue paper roll,\\ trash bag, trashcan, tumbler, wastebin, tissue dispenser stand, trash can, coffee, duster, dusting cloth, dustpan,\\ handle, iron, knob, drain, floor wiper\end{tabular} \\
Tables & conference table, dining table, folded table, coffee table, desk, short table, sidetable, study table, joined tables \\
Seating & \begin{tabular}[c]{@{}l@{}}arm chair, barber chair, chair, chairs, couch, dining chair, easy chair, footrest, footstool, l-shaped sofa, folding sofa,\\ floor couch, floor sofa, foot rest, lounge chair, office chair, office visitor chair, rolling chair, seat, sofa, sofa chair, stool,\\ stools, upholstered bench\end{tabular} \\
Bedroom & bed, bedframe, bedpost, headboard, duvet, bedsheet, bed sheet, bed cover, mattress \\
Pillows \& Cushions & pillow, chair cushion, cushion, cushions, long pillow, beanbag, seat cushion, sit-up pillow, sofa cushion \\
Fabrics \& Mats & cloth, cloth piece, canvas, blanket, fabric, gym mattress, leather mattress, rolling mat, sheet, sheets, yoga mat \\
Storage \& Shelves & \begin{tabular}[c]{@{}l@{}}bedside cabinet, bedside counter, wardrobe, cabinet, cabinet side panel, cabinet top, closet rail, garage shelf, \\ glass shelf, fume cupboard, clothes cabinet, drawer, fitted wardrobe, laundry hamper, nightstand, organizer, \\ plastic container, recessed shelf, shelf, small cabinet, switchboard cabinet\end{tabular} \\
Bags & backpack, packet, plastic bag, pouch, rubber water bag, sack, shopping bag, wicker basket \\
Background & \begin{tabular}[c]{@{}l@{}}carpet, baseboard, billboard, brick wall, ceiling, column, floor, floor mat, floor mats, rolled backdrop,\\ fake ceiling, glass, glass pane, glass wall, mount, panel, partition wall, pillar, plank, backdrop, ceiling pipe, \\ ceiling beam, roof, sill, steel beam, suspended ceiling, valve, vent, air vent, ceiling vent, exhaust duct, \\ exhaust pipe, wall, wall beam, wall board, wall cord cover, wall strip, duct\end{tabular} \\
Doors \& Windows & \begin{tabular}[c]{@{}l@{}}curtain, curtain rail, blinds rail, blinds rod, roller blinds, rolling blinds, blinds, skylight, vertical blind control,\\  vertical blinds, window sill, windowframe, door frame, door window, cubicle door, cabinet door, door, doorframe\end{tabular} \\
Decor & \begin{tabular}[c]{@{}l@{}}decorative mirror, decorative object, flag, painting, sculpture, vase, plant pot, candle holder, clock, coat stand,\\ dried plant, full-length mirror, mirror frame, photo, picture, picture frame, plant, plant pot coaster, plushie,\\ potted plant, statue, table clock\end{tabular} \\
Lighting \& Lamps & \begin{tabular}[c]{@{}l@{}}bedside lamp, ceiling lamp, desk light, lamp, fluorescent lamp, light, light cover, light fixture, light panel, \\ light switches, softbox, softbox light, spotlight, table lamp, wall light\end{tabular} \\
Office & \begin{tabular}[c]{@{}l@{}}file binder, file cabinet, file holder, file organizer, file stack, file tray, folder, folder organizer, magazine holder,\\ paper tray, book, books, document holder, interactive board, interactive whiteboard, eraser, files, marker eraser,\\ palm rest, paper, paper roll, pencil, pencil case, post it, blackboard, white board, whiteboard mount, stapler, \\ paper stapler, scissor\end{tabular} \\
Electronics & \begin{tabular}[c]{@{}l@{}}electronic device, laptop, in-table power socket, computer tower, card reader, mobile tv stand, monitor, monitor cover,\\ monitor light, power sockets, powerstrip, projector screen, scanner, screen, socket, speaker, switch, tablet, tv screen, \\ web cam, webcam, rolled projection screen, cable tray, heater, covered power socket\end{tabular} \\
Clothing \& Accessories & \begin{tabular}[c]{@{}l@{}}bathrobe, belt, cap, coat, glasses case, glasses cover, hoodie, jacket, logo, pants, shirt, shoe, suit, suit cover, sweater,\\ t shirt, trousers, hair brush, comb\end{tabular} \\
Bathroom & bathroom mat, bathroom stall, bathroom shelf, toilet flush button, toothpaste, towel, tub \\
Misc & \begin{tabular}[c]{@{}l@{}}anti splatter shield, cat bed, paint jar, cylinder, beam, button, cord, high pressure cylinder, \\ cooling pad, leaf fan, machine button, pipe, product tube, razor blade, rod, rods, silicon gun, \\ spool, squeegee, stick, ar tag, ball, cream tube, barrel, brief, cardboard, handrail, metal mount, \\ metal sheet, prosthetic leg, reflection, shade, stamp, sticker, umbrella, wood panel, wood piece, wood stick\end{tabular} \\ \bottomrule
\end{tabular}%
}
\label{tab:categories}
\end{table*}
 
\begin{figure*}[ht]
    \centering
    \includegraphics[width=\textwidth]{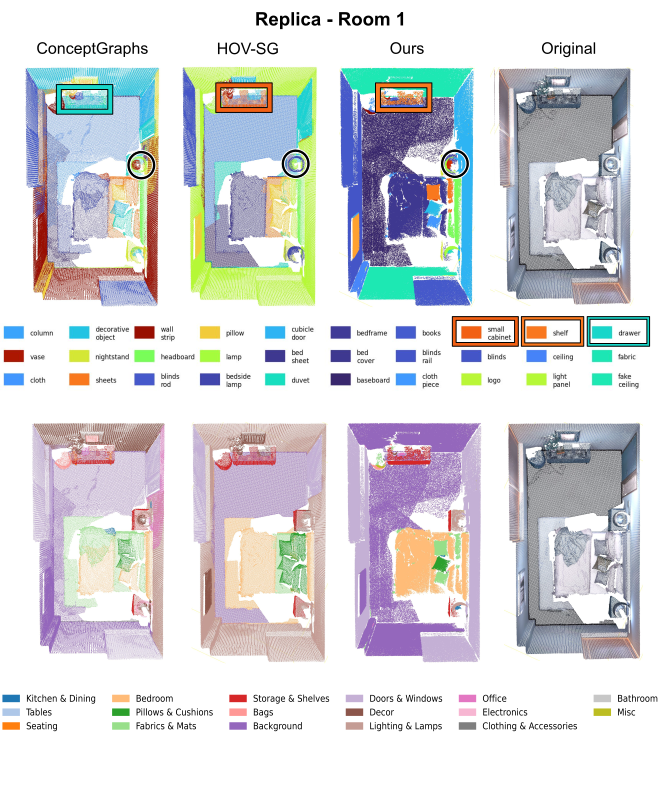}
    \label{fig:pcds}
\end{figure*}

\begin{figure*}[ht]
    \centering
    \includegraphics[width=\textwidth]{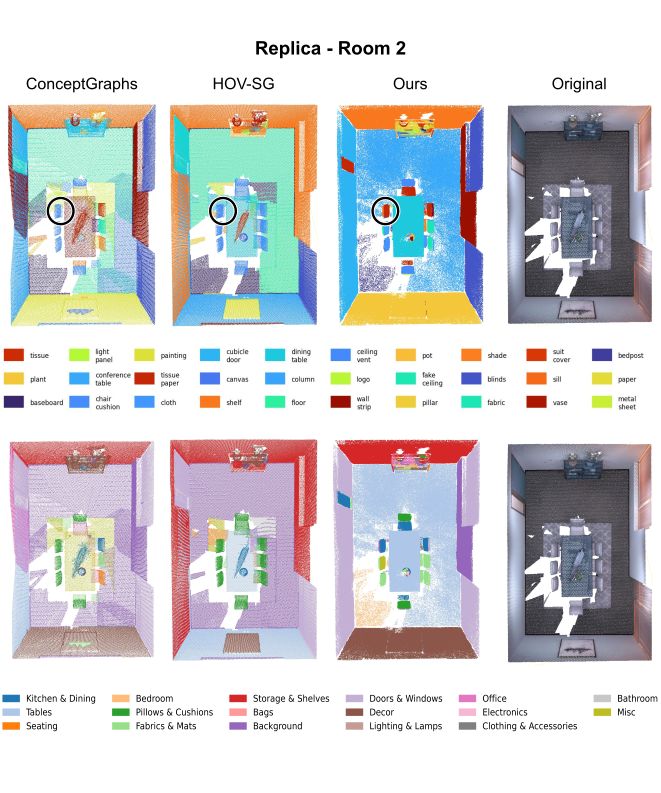}
\end{figure*}

\begin{figure*}[ht]
    \centering
    \includegraphics[width=\textwidth]{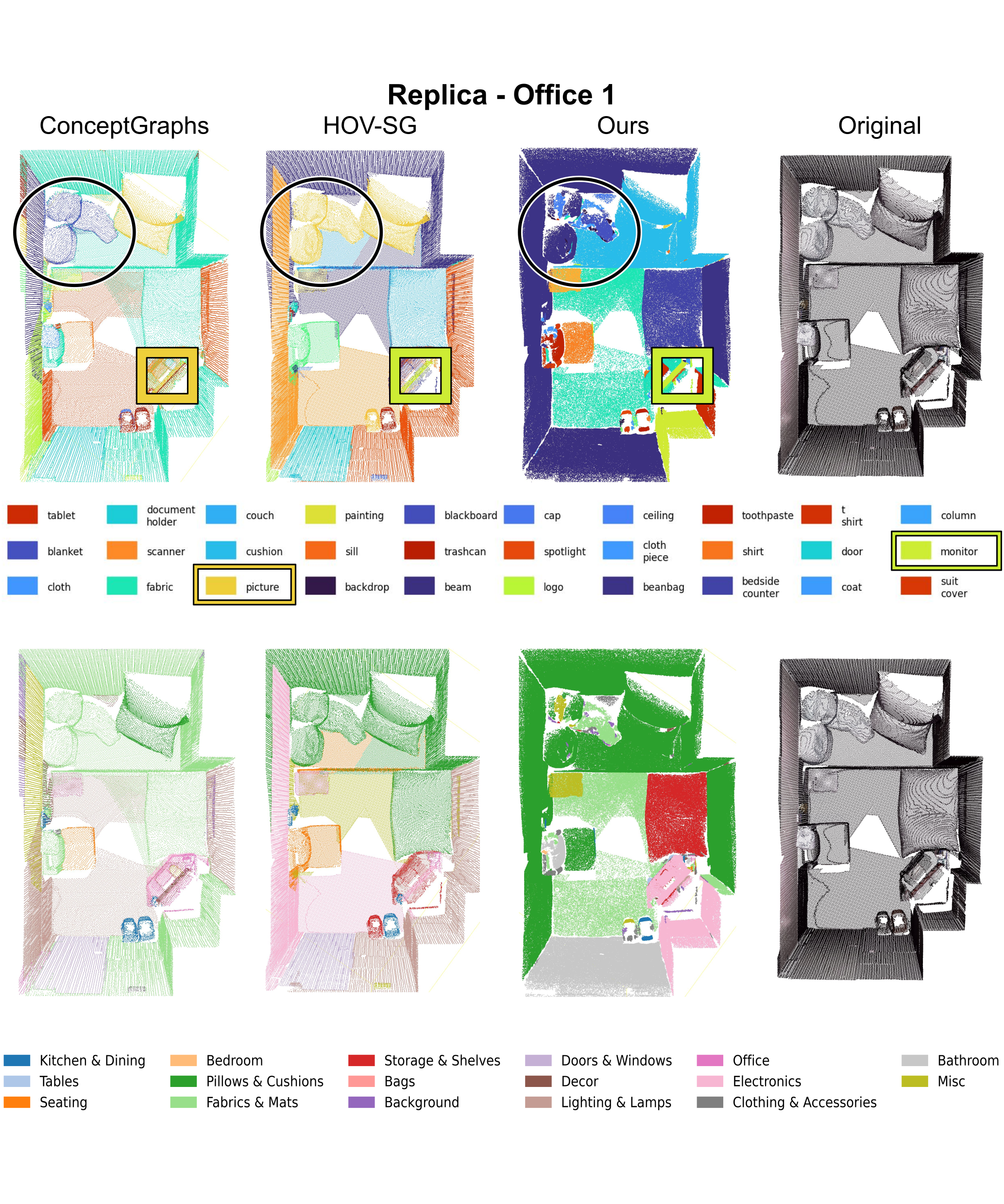}
\end{figure*}

\begin{figure*}[ht]
    \centering
    \includegraphics[width=\textwidth]{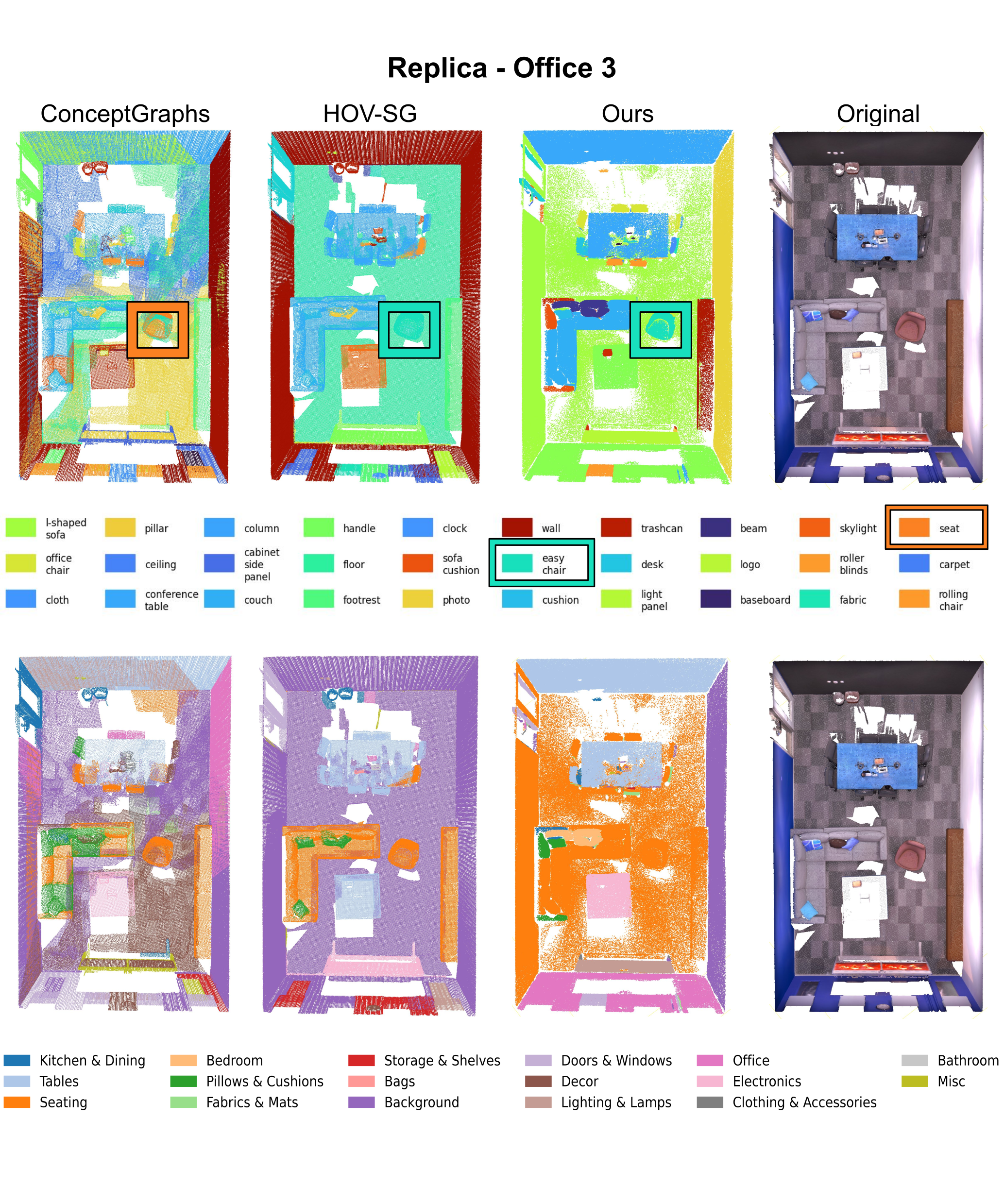}
\end{figure*}

\end{document}